\documentclass[preprint,12pt]{elsarticle}
\usepackage[english]{babel}
\usepackage{graphicx}
\usepackage{float}
\usepackage{amssymb}
\usepackage{amsthm}

\journal{XXXX}

\begin{document}

\begin{frontmatter}

\title{Edge-set reduction to efficiently solve the graph partitioning problem with the genetic algorithm}

\author[]{Ali CHAOUCHE}
\author[]{Menouar BOULIF}
\address{LIMOSE Laboratory \\
Department of Computer Science, University Mhamed Bougara of Boumerdes, Avenue de l’independance, 35000. Boumerdes, Algeria}

\begin{abstract}
The graph partitioning problem (GPP) is among the most challenging models in optimization. Because of its NP-hardness, the researchers directed their interest towards approximate methods such as the genetic algorithms (GA). The edge-based GA has shown promising results when solving GPP. However, for big dense instances, the size of the encoding representation becomes too huge and affects GA's efficiency. \\
In this paper, we investigate the impact  of modifying the size of the chromosomes on the edge based GA by reducing the GPP edge set. We study the GA performance with different levels of reductions, and we report the obtained results. 

\end{abstract}

\begin{keyword}
 Graph partitioning problem \sep Genetic algorithm \sep Encoding representation \sep Edge-based encoding \sep Edge set reduction.

\end{keyword}

\end{frontmatter}

\section{Introduction}
\label{Intro}
The graph partitioning problem (GPP) is one of the combinatorial optimization problems whose applicability spectrum covers a huge number of quite dissimilar domains \cite{chaouche2019solving}. \\
GPP seeks to partition the vertices of a graph into moderately-sized disjoint subsets so as to optimize the cost of the overall cut weight. GPP ubiquitousness stems from a key factor. Indeed, GPP can be used to materialize the "divide to conquer" philosophy, which by decomposing complex systems into less complicated subsystems eases their handling in order to solve the entire problem.  Therefore, GPP has been used in many areas including parallel computing \cite{murni2017hypergraph}, power network design \cite{li2010controlled}, VLSI design \cite{kahng2011vlsi}, image processing \cite{peng2013survey}, road networks \cite{luxen2012candidate}, etc. \\
Being NP-Complete, the use of exact methods to solve GPP was not considered as a first concern by the research community. As a consequence, the use of approximate methods such as genetic algorithms (GA) seems to be the most promising path. The performance of the GA is closely related to the encoding scheme used to represent the solutions of the problem. Therefore, several encodings have been proposed in the literature \cite{menouar_2010, chaouche2019solving} such as the integer encoding \cite{venugopal_narendran_1992}, the fractional representation  \cite{gonccalves2002hybrid}, the edge based encoding (EE) \cite{boulif_atif_2006}, etc. This latter presents a set of promising features such as reduced alphabet, good redundancy, etc. \cite{menouar_2010}. EE representation, in contrast to the widespread used encodings, doesn't use vertex but edge assignment to define a graph partition. Although the associated reported results were promising, the authors noticed that the performance varied from one instance to another. In this paper we trial the assertion that such a behaviour is essentially governed by the density of the graph. That is, EE performs well when this density is moderate.  Hence, when a graph has an important density, we can take out some edges from being considered in the encoding representation.   
The rest of this work is organized as follows: in section 2 a formal description of the graph partitioning problem is given. The third section is devoted to the presentation of genetic algorithms and their principles. Section 4 outlines the  EE representation. The next section explains the reduction approach. In section 6, we present the results of the empirical analysis. Finally, in the last section, we draw some conclusions.

\section{Formal definition of the graph partitioning problem}
\label{formalGpp}
We consider an undirected graph $G=(V,E)$ such that $V = \{v_1,v_2,\dots, v_{|V|}\}$ is the set of vertices, and $E = \{e_1,e_2,\dots, e_{|E|}\}$ is the set of edges. Each edge $e \in E$ has a positive weight $\omega(e)$. Partitioning $G$ consists of finding a partition $P = \{C_1,C_2,\dots, C_{|P|}\}$ of $V$ into $|P|$ disjoint clusters, i.e .:
\begin{equation}
	\forall i,j \in \{1,2,\ldots,|P|\},i\neq j : C_i\cap C_j= \emptyset
\end{equation}

\begin{equation}
	C_1 \cup C_2 \cup \dots C_{|P|} = V 
\end{equation}
The objective is to minimize the total cut size. That is,
\begin{equation}
 \sum_{e_{ij} \in \{ \{u,v\} \in E : u \in C_i, v \in C_j, ,i\neq j \}}\omega(e_{ij}) 
\end{equation}
In order to be  accepted, the partition must satisfy the following constraints:
\begin{itemize}
	\item Moderate cluster size : the number of vertices per cluster has an upper bound.
	\item Cohabitation constraints: some vertices must be put in the same cluster.
	\item Non cohabitation constraints: some vertices must be put in different clusters.
\end{itemize}

\section{Genetic algorithms}
\label{GA}
With his publication "Adaptation in  Natural  and  Artificial  Systems" \cite{holland_1975} John Holland is considered the father of the genetic algorithms. After that, David Goldberg came in 1989 to vulgarize their concepts \cite{goldberg1989genetic}. Since then, GAs enjoyed much attention due to their ability to solve complex problems. Genetic algorithms imitate the principle of evolution of natural species by evolving a set of chromosomes (individuals) that represent solutions of the problem to be solved. Two primary processes are used to ensure this evolution: natural selection and sexual reproduction. The first determines the most fitted individuals to their environment to survive and reproduce, and the second ensures mixing and recombination of the individuals to make offspring.
To solve optimization problems using GAs, we proceed as follow : first an encoding scheme is defined to represent candidate solutions (the individuals of the population). After that, each individual in the population is evaluated according to the objective function. Then, the higher-ranking individuals mate to produce a new generation. With a low probability, the offspring can be modified by a mutation operator. This process is iterated until a certain stopping criterion is reached. Among the genetic representations, we use the edge based encoding for which the following section is devoted.

\section{Edge based representation}
\label{BSEB}
\begin{figure}[H]
	\centering
	\includegraphics[width=0.6\linewidth]{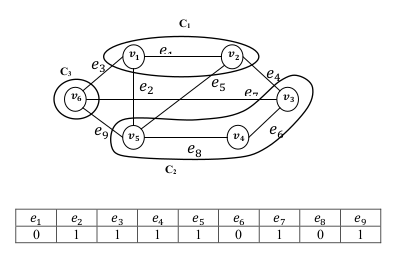}
	\caption{An example of a graph partition of three clusters with its EE representation.}
	\label{fig1}
\end{figure}

In their work \cite{boulif_atif_2006}, the authors proposed a binary edge based encoding scheme to represent the individuals of the population (chromosomes) in order to solve the  GPP. The principle of this encoding scheme is given as follow : 
\begin{itemize}
	\item \textbf{Step 1.} The edges are lexicographically sorted according to their incident vertices. See for example how the edges are indexed in Fig. \ref{fig1}.
	\item  \textbf{Step 2.} Each individual of the population are then represented by a binary chain of length $|E|$ where each component corresponds to an edge. The corresponding allele is either 0 for an intra-cluster (the two extremities of the edge are in the same cluster) or 1 for an inter-cluster edge (the extremities are in different clusters).
\end{itemize}
In the example of figure \ref{fig1},  the intra-cluster edges $e_1=\{v_1,v_2 \}$, $e_6=\{v_3,v_4 \}$, $e_8=\{v_4,v_5\}$ form the first and the second cluster of the partition, the third cluster is formed only by $v_6$. The rest of edges are inter-clusters and the sum of their weights gives the cut size of the partition.

\section{Edge set reduction}
\label{weighThreshold}
\begin{figure}[H]
	\centering
	\includegraphics[width=0.8\linewidth]{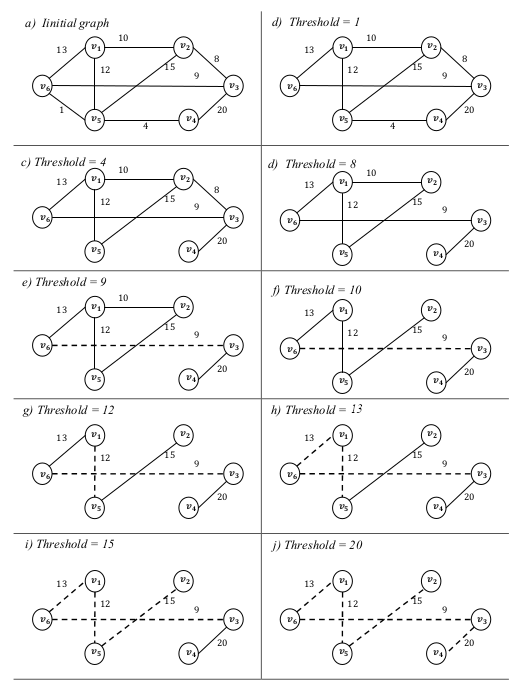}
	\caption{An example of  the application of a threshold on the graph edges.}
	\label{fig2}
\end{figure}

The EE representation proposed in \cite{boulif_atif_2006} uses the overall set of edges to represent solutions. Unfortunately, this definition may hinder the performance of the GA, especially for dense graph instances, that is, when there is a  considerable number of edges. In fact, the representation of a partition using this encoding scheme does not need to use the entire set of edges, but only a subset that yields a partial graph. 
For this purpose, we consider a threshold $T$ on the edge weights to be considered for the construction of the solutions. Every edge whose weight is lesser that $T$ will not take part of the representation. Then, by varying $T$, we can analyse the impact on the GA performance to find the most suitable threshold.
To make thigs more clear, let us take for example the graph instance of figure \ref{fig2}.a. Applying different threshold values yields the graphs of figure \ref{fig2}.$d$ to $j$. 

\noindent The weight thresholding technique removes all the edges with a value that is lower than or equal to the threshold. However, if the removal of an edge affects the graph connectivity, the edge is kept back. The following algorithm describes the  edge reduction procedure.

\begin{itemize}
	\item \textbf{Step1:} Delete all the edges whose weight is lower than or equal to the threshold. 
	\item \textbf{Step2:} Check the connectivity of the graph resulting from Step 1. If it is not connected, restore the removed edge with the greatest weight. An edge is restored only if the number of connected components decreases. This process is repeated until the graph becomes connected.
\end{itemize}

\noindent In figure \ref{fig2}, dashed lines represent edges with weights lower than or equal to the threshold that have been restored because their removal disconnected the graph.

\section{Results and analysis}
\label{resAnalysis}

In order to perform a fast analysis, we have used the same parameters when applying the GA to all the graph instances. Table \ref{GA_platform} summarizes the associated values and descriptions.

\begin{table}[h]
	\begin{center}
	\begin{tabular}{ll}
		\hline
		\textbf{Parameter} & \textbf{Description}\\
		\hline
		Selection &	90\textdiscount \\

		Elitism	& 10\textdiscount \\

		Crossover rate & 80$\%$ \\
		Mutation rate &	2$\%$ \\
		Population size	& 100 \\
		Number of generations &	100 \\
		Number of runs &	30 \\
		\hline
	\end{tabular}
\end{center}
	\caption{Genetic algorithm parameters.}
	\label{GA_platform}
\end{table}
To measure the performances of the GA, we have used three metrics. The first one is the Average Best Fitness over thirty runs (ABF), used to evaluate the efficiency of the GA. The Number of the Visited Solutions (NVS) metric is used to estimate the effort made by the algorithm to reach its best solution. The speed of the AG is measured using the Mean Best Run Time (MRT). 
Among the benchmarks we used in our experiments, we have chosen three instances to show the impact of our approach on the performances of the EE GA. Table \ref{GraphInstances} illustrates the features of the benchmarks.

\begin{table}[h]
	\begin{center}
		\begin{tabular}{ccccc}
			\hline
			\textbf{Graph $\#$} & \textbf{Number of edges} & \textbf{Number of vertices} \\
			\hline
			 1 & 55 & 15 \\
			 2 & 45 & 30 \\
			 3 & 1980 & 121 \\
			\hline
		\end{tabular}
	\end{center}
	\caption{Graph instances features.}
	\label{GraphInstances}
\end{table}

\begin{figure}[H]
	\centering
	\includegraphics[width=0.9\linewidth]{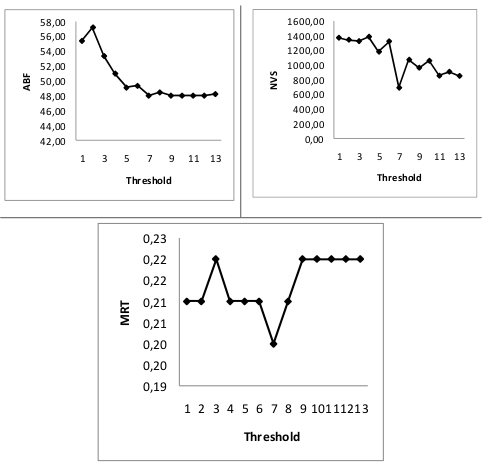}
	\caption{Results of  the first instance.}
	\label{result1}
\end{figure}

\begin{figure}[H]
	\centering
	\includegraphics[width=0.9\linewidth]{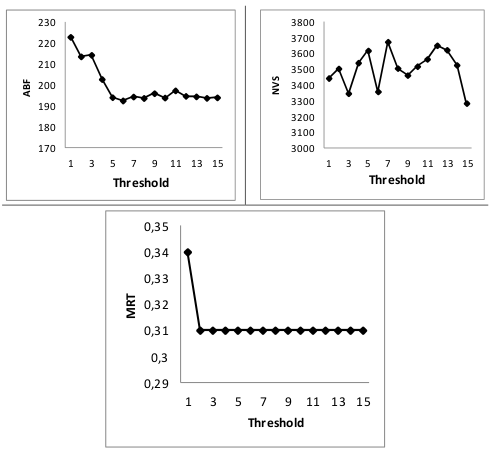}
	\caption{Results of  the second instance.}
	\label{result2}
\end{figure}

\begin{figure}[H]
	\centering
	\includegraphics[width=0.9\linewidth]{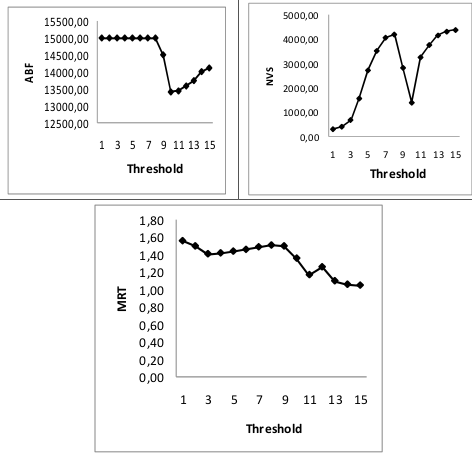}
	\caption{Results of  the third instance.}
	\label{result3}
\end{figure}

The results of our experiments confirm the impact of the edge set modification on both GA effectiveness (solution quality) and efficiency (algorithm speed).  However,  there is no straightforward  formula that determines the exact number of the removed edges since taking away a large amount of edges hinders the efficiency of the GA too. Indeed,  the behavior of the GA changes from an instance to another. In fact,  removing certain edges can make the edge EE encoding incapable to represent some solutions of the search space. This situation becomes more critical if the encoding scheme cannot represent optimal solutions \cite{boulif_atif_2006,menouar_2010}. For example, the optimal solution for the graph of figure 6 cannot be encoded with the EE representation for a threshold equal to 4.

\begin{figure}[H]
	\centering
	\includegraphics[width=1\linewidth]{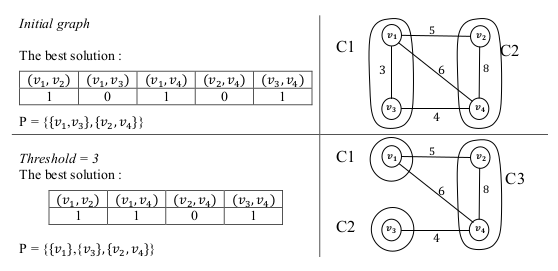}
	\caption{The best solutions visible by the encoding scheme with threshold equal to 4.}
	\label{result4}
\end{figure}

On the other hand, increasing the threshold does not always mean decreasing the number of edges in the graph. For example, if we set the threshold to 4, we get the same graph, because removing the edge ${v_3,v_4}$ disconnects the graph.
Therefore, the performances of the GA using the EE representation are boosted for a compromise threshold for which the GA reaches its peak in terms of efficiency. Nevertheless, the value of this threshold seems not to be only linked to the number of edges. The experimental results show that not only the number of edges but their weights influences the definition of the sought threshold. This can be confirmed by Fig. \ref{result3}. Indeed, for a threshold equal to 10 the GA reaches its best score in terms of ABF, whereas for a very small number of edges the GA gives a poor performance.

\section{Conclusion}
\label{conclusion}
In this work we propose an enhancement for the edge- encoding representation in the genetic algorithm to solve the graph partitioning problem, for graphs with quite dissimilar edge weights. In order to enhance the efficiency of the GA, a threshold value enables to reduce the number of edges to be considered in representing the solutions.\\
By using different threshold values, the experimental results show that the GA efficiency is not only governed by the number of edges, but also by how their weights are spread throughout the graph. Further experimental investigations are needed to directly estimate the compromise threshold value based on the edge-weight underneath structure of the input graph. For future work, we are interested by investigating how the thresholding approach impacts the dynamic version of the graph partitioning problem.

\bibliographystyle{elsarticle-harv} 
\bibliography{referencesBib}

\end{document}